\documentclass{article}
\usepackage[table,xcdraw]{xcolor}
\usepackage{ijcai23}
\usepackage{multirow}
\usepackage{times}
\usepackage{soul}
\usepackage{url}
\usepackage[hidelinks]{hyperref}
\usepackage[utf8]{inputenc}
\usepackage[small]{caption}
\usepackage{graphicx}
\usepackage{amssymb}
\usepackage{amsthm}
\usepackage{booktabs}
\usepackage[switch]{lineno}
\usepackage{amsfonts}
\usepackage{algorithm}
\usepackage{algpseudocode}
\usepackage{amsmath}

\title{Self-Supervised Neuron Segmentation with Multi-Agent Reinforcement Learning}

\author{
Yinda Chen$^{1,2}$
\and
Wei Huang$^1$\and
Shenglong Zhou$^1$\and
Qi Chen$^1$ \and
Zhiwei Xiong$^{1,2,}$\footnote{Corresponding Author}
\affiliations
$^1$University of Science and Technology of China\\
$^2$Institute of Artificial Intelligence, Hefei Comprehensive National Science Center
\emails
\{cyd0806, weih527, slzhou96, qic\}@mail.ustc.edu.cn,
zwxiong@ustc.edu.cn
}

\begin{document}

\maketitle

\begin{abstract}
The performance of existing supervised neuron segmentation methods is highly dependent on the number of accurate annotations, especially when applied to large scale electron microscopy (EM) data. By extracting semantic information from unlabeled data, self-supervised methods can improve the performance of downstream tasks, among which the mask image model (MIM) has been widely used due to its simplicity and effectiveness in recovering original information from masked images. However, due to the high degree of structural locality in EM images, as well as the existence of considerable noise, many voxels contain little discriminative information, making MIM pretraining inefficient on the neuron segmentation task. To overcome this challenge, we propose a decision-based MIM that utilizes reinforcement learning (RL) to automatically search for optimal image masking ratio and masking strategy. Due to the vast exploration space, using single-agent RL for voxel prediction is impractical. Therefore, we treat each input patch as an agent with a shared behavior policy, allowing for multi-agent collaboration. Furthermore, this multi-agent model can capture dependencies between voxels, which is beneficial for the downstream segmentation task. Experiments conducted on representative EM datasets demonstrate that our approach has a significant advantage over alternative self-supervised methods on the task of neuron segmentation. Code is available at \url{https://github.com/ydchen0806/dbMiM}.

\end{abstract}

\section{Introduction}
\begin{figure}[t]
    \centering
    \includegraphics[width = \linewidth]{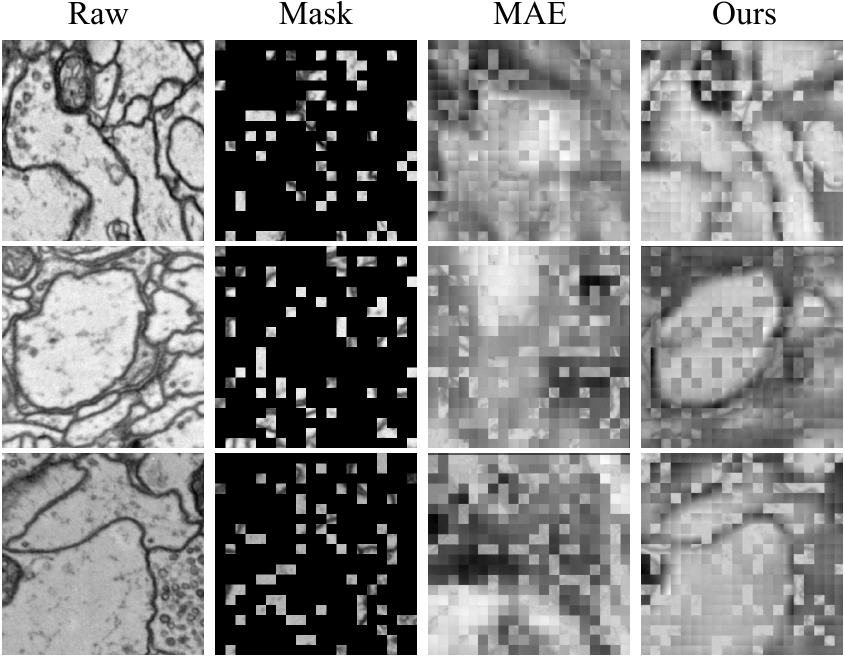}
    \caption{A comparison of the reconstruction effectiveness of our proposed method with MAE. The first column shows the original EM image, the second column shows the image after masking 85\% of the voxels, the third column shows the reconstruction using the MAE method, and the fourth column shows the reconstruction using our method.}
    \label{reconstruction}
    
\end{figure}
Neuron segmentation is a crucial task for neuroscientists that allows for the analysis of the distribution and morphology of neurons, providing valuable insights into the connectomics research \cite{sheridan2022local,krasowski2017neuron}. Electron microscopy (EM) is the mainstream method for accurately identifying neural structures, but the dense nature of neurons and the presence of artifacts and deformations in EM images make the labeling process costly and decrease the credibility of existing annotation data \cite{deng2022unified,zhou2019fast,chen2023learning}. Therefore, fully supervised neuron segmentation methods meet great challenges, especially when applied to large scale EM data.

Self-supervised methods have emerged as a solution to the limitations of fully supervised methods, which can be roughly divided into two categories: contrastive learning-based approach and mask image model (MIM)-based approach. The former requires a large number of positive and negative samples \cite{chen2020simple,grill2020bootstrap,you2022simcvd,chen2023generative} and relies heavily on data augmentation \cite{caron2021emerging}, making it a high-cost option for 3D biomedical images. The latter aims to learn useful structural information in images by masking and recovering certain voxels, which has been recently applied to pretraining biomedical images, showing improvements in downstream tasks \cite{zhou2022self,tang2022self,huang2022semi}. 
However, the highly localized and structured nature of EM data, as well as the existence of considerable noise, make it inefficient to directly employ the existing MIM in extracting useful information for neuron segmentation. It has also been observed that the masking ratio and masking strategy of MIM are highly sensitive and the optimal ones vary greatly across different datasets. Adjusting these configurations to train large models requires significant manual efforts and resources.

In this paper, targeting the neuron segmentation task, we propose a novel decision-based MIM relying on multi-agent reinforcement learning (MARL) \cite{littman1994markov} for automatically selecting the appropriate masking ratio and masking strategy, which consists of a target network and a policy network. Our approach partitions the input EM volume into patches and treats each patch as a basic control unit. The overall multi-agent task is modeled as a search process for patch masking strategies, where the action space for each patch is to either keep the original voxels or mask them. The feedback of the target network, in the form of the reconstruction loss, serves as the team reward signal for guiding the policy network to adaptively learn masking strategies that are beneficial for the pretraining task \cite{foerster2016learning}. To improve the stability of training and achieve optimal joint decision-making for the entire volume, all agent networks share parameters and are trained in parallel. Furthermore, we introduce the HOG feature as an additional reconstruction loss to enable the target network to learn more structure information. Finally, following a UNETR decoder design \cite{hatamizadeh2022unetr}, we add a segmentation head to the pretrained target network in the finetuning stage. Experimental results in Figure \ref{reconstruction} show that our decision-based MIM achieves clearer reconstruction results than the original MAE \cite{he2022masked} in the pretraining phase.

Overall, our main contribution lies in the following aspects:

1) We propose an efficient self-supervised method, named decision-based MIM, for EM neuron segmentation using unlabeled EM data. To the best of our knowledge, it is the first effort that large-scale transformer pretraining is conducted on this task. 

2) We propose a MARL-based approach for searching the optimal masking ratio and masking strategy by treating each patch as an agent with a shared policy, effectively exploring the search space and capturing dependencies between voxels.

3) We introduce the HOG feature as an additional reconstruction loss of our decision-based MIM, improving the convergence speed of network training and the performance of the downstream segmentation task. 

4) We comprehensively demonstrate the effectiveness of our proposed method on two representative EM datasets, especially against alternative self-supervised methods on the task of neuron segmentation.

\section{Related Work}
\subsection{Neuron Instance Segmentation}
In the field of EM image processing, neuron instance segmentation is an important task. \cite{turaga2010convolutional} first proposed a convolutional neural network based on affinity generation, followed by clustering voxels in the affinity graph into instances based on post-processing methods such as watershed and LMC \cite{beier2017multicut}. In recent years, there have been more advanced networks for the affinity-based approach. Funke et al.\cite{funke2018large} introduced the MALIS loss\cite{briggman2009maximin} during the training process to encourage the network to generate correct topological segmentation. \cite{huang2022learning} introduced an embedding pyramid module to simulate affinity at different scales. \cite{liu2022biological} incorporated both embedding and affinity information and combined it with the graph neural network to further improve the distinguishability of adjacent objects in the feature space. However, due to the anisotropic resolution of 3D EM images in lateral and axial directions, the usage of transformer structures remains unexplored in the field of neuron segmentation. In this paper, we use an affinity-based setup and upgrade the decoder of UNETR \cite{hatamizadeh2022unetr} to adapt to the anisotropic EM features.

\begin{figure*}[ht]
    \centering
    \includegraphics[width = 0.8\linewidth]{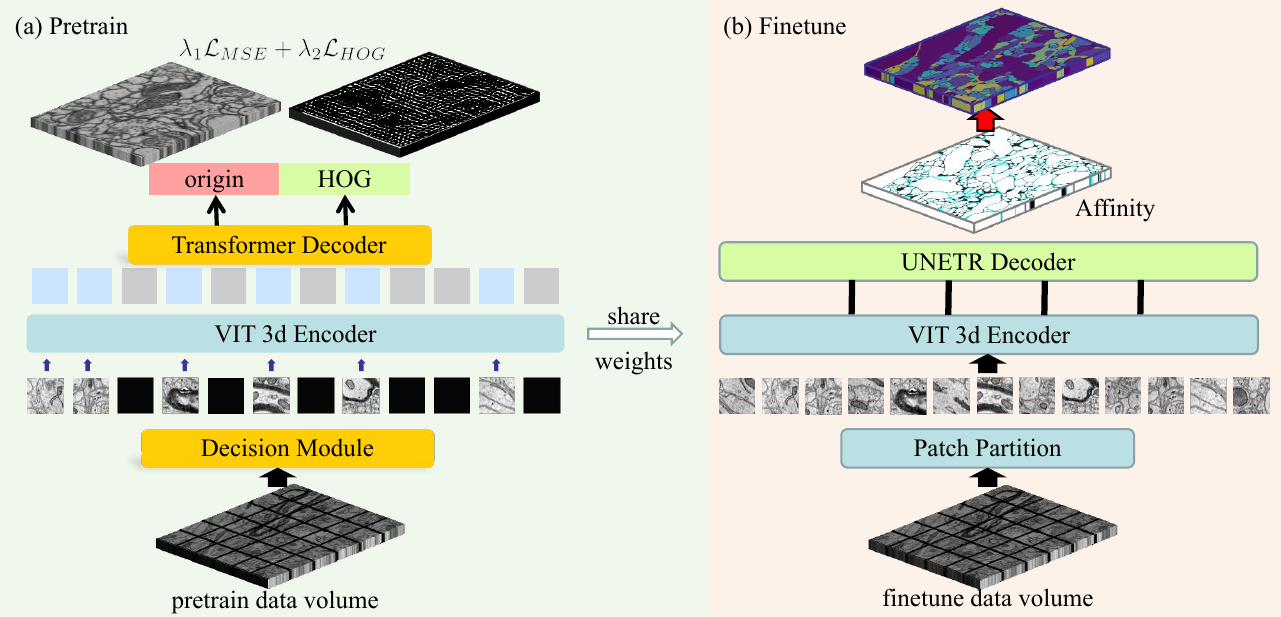}
    \caption{Our proposed network architecture is divided into two main components: (a) the decision-based Mask Image Model (MIM) pretraining process, which employs our proposed decision module to select appropriate patches for masking and then utilizes a 3D Vision Transformer (ViT) encoder to encode the visible patches. The resulting tokens are then passed through a lightweight decoder to reconstruct the original voxels and histograms of oriented gradient (HOG) features. (b) The fine-tuning process for the downstream segmentation task utilizes the encoder weights from the pretraining process and adds a UNETR segmentation head to output the affinity map. The final segmentation results are obtained through post-processing methods such as waterz.}
    \label{network}
\end{figure*}

\subsection{Mask Image Model (MIM)}
The MIM is an important branch of self-supervised learning. Masked autoencoders (MAE) \cite{he2022masked} used an asymmetric encoding-decoding structure, encoding only the unmasked patches and using a lightweight decoder to recover the masked patches, which greatly reduced the resources required for computation and quickly became a mainstream structure for MIM. \cite{feichtenhofer2022masked} and \cite{tong2022videomae} separately validated the effectiveness of MAE on video datasets and proved that higher image masking ratios can be used in 3D datasets. \cite{zhou2022self} was the first to introduce MAE to the medical image field. \cite{bachmann2022multimae} introduced a multi-modal multi-task adaptation to MAE, resulting in better performance than the original MAE. \cite{gaomcmae} proposed a multi-scale mixed convolution to encode images and achieved improved results on fine-grained downstream tasks such as image segmentation. \cite{wei2022masked} focused on the reconstruction target of the decoder and found that reconstructing artificial features such as HOG and SIFT can facilitate the network to classify and localize objects. 

In summary, existing MIM-based methods mainly focus on the design of the encoder architecture, prediction head, and prediction target. Although many works have demonstrated the impact of masking strategies on downstream tasks, there has been little research along this line, and the latest method also requires a search-based masking strategy with a fixed masking ratio \cite{bandara2022adamae}. This paper proposes a novel MARL-based approach for adaptively learning the optimal masking ratio and masking strategy, making the pretrained model more robust and achieving better performance on downstream tasks.

\subsection{Multi-Agent Reinforcement Learning (MARL)}
In the field of deep RL, multiple agents working together can improve the efficiency and robustness of the model due to the limited observation and action space of a single agent. Given the complexity of computer vision tasks,  MARL is often used to interact with the common environment to make decisions. \cite{liao2020iteratively} proposed a method for iteratively refining medical image segmentation using interactive MARL, where rough image segmentation is initially provided and the network is iteratively refined through user feedback in the form of rewards until the segmentation is sufficient. \cite{lin2021local} proposed a method for augmenting images through the use of blocks, with each block acting as an agent and working together to produce optimal data augmentation. Still, the application of MARL in computer vision suffers from the large decision space and difficulty in obtaining rewards. In this paper, to reduce the complexity of the state and the difficulty of searching for RL policies, we first segment the image into patches through a transformer-encoder and treat each patch as an agent. The action space for each patch is limited to only two options, masking or keeping the original voxels, and the rewards are obtained through the reconstruction loss of MAE. Therefore, the masking decision can be naturally modeled as a Markov process and optimized through MARL.

\section{Proposed Method}

Our decision-based MIM consists of two stages of training: pretraining and finetuning. Figure \ref{network} illustrates the overall flow of the network training. In this section, we will first introduce some basic theories of Vision Transformer (ViT) and MARL, and then explain in detail the encoders, decoders, and loss functions of the two stages, as well as specific MARL modeling methods.

\subsection{Encoder-Decoder Design}
We use ViT \cite{dosovitskiyimage} as the backbone architecture for decision-based MIM pretraining and downstream segmentation tasks. To represent high-dimensional data in a ViT, we must transform it into a sequence of patches. Given an input 3D volume $\mathbf{x} \in \mathbb{R}^{H \times W \times D \times C}$, where $C$ is the number of channels and $(H, W, D)$ is the resolution, we reshape it into a sequence of flattened 3D patches $\mathbf{x}_p \in \mathbb{R}^{N \times (P^3 \cdot C)}$. The patch resolution is given by $(P/4, P, P)$, and the number of patches is calculated as $N = \frac{4HWD}{P^3}$. These patches are then transformed into patch embeddings via a trainable linear projection.

Consistent with the MAE setup, we divide the image patches into visible and masked groups. The encoder in the MAE and ViT architectures processes only the visible blocks. To enhance the performance of our decoder, we incorporated a histogram-of-oriented-gradients (HOG) feature \cite{dalal2005histograms}, which has been shown to improve pretraining. This is achieved by providing the decoder with various markers, including a patch representation from the encoder and learnable position embeddings. By incorporating positional embedding in all input markers, we enable the decoder to simultaneously recover both the HOG feature and the original voxels, resulting in superior performance compared to the MAE method.

In our pretraining process, we utilize a loss function that combines both the mean squared error (MSE) loss for reconstructing the original voxels and the HOG loss for recovering the HOG features. The HOG feature can be calculated using the following equation
\begin{equation}
    \text{HOG}_{i,j}=\frac{\sum_{x\in S_{i,j}}{w(x)g(x)}}{\sum_{x\in S_{i,j}}{w(x)}},
    \label{HOG}
\end{equation}
where HOG$_{i,j}$ is the histogram of oriented gradients for the cell located at position $(i,j)$, $S_{i,j}$ is the set of voxels in the cell $(i,j)$, $w(x)$ is a weighting function that assigns a weight to each voxel $x$ in the cell, and $g(x)$ is the gradient orientation of the voxel $x$.

Our overall loss function can be expressed as
\begin{equation}
\mathcal{L}_{pretrain} = \lambda_1 \mathcal{L}_{M S E}+\lambda_2 \mathcal{L}_{H O G},
\label{loss_pre}
\end{equation}
where $\lambda_1$ and $\lambda_2$ denote the weights assigned to the MSE and HOG losses, respectively. We set $\lambda_1$ to 0.1 and $\lambda_2$ to 1.
\subsection{Decision Module}
In our proposed decision module, we model the masking strategy of image patches in MAE as a multi-agent cooperative decision-making problem and adopt a multi-agent reinforcement learning method to solve it. Here, we will introduce in detail the modeling methods of model states, observations, actions, the design of multi-agent team rewards, and the learning method used to update the policy network.

As shown in Figure \ref{MARL}, given the original input batch of image $\mathbf{x}$, according to the ViT setup, we divide the image into equal-sized and non-overlapping patches. Our MARL policy aims to determine the overall joint masking policy based on the current input state and the observation of each agent. Our basic setup is as follows.

\subsubsection{State.} The enhancement policy for each patch is closely related to the information of the context, so the decision-making process of MARL requires perceiving the semantic information of the image rather than directly inputting the patches of the raw image. In order to ensure the consistency and convergence of training, we use the target network, i.e. ViT, as the backbone to extract deep semantic features of the image. The global state at time step $t$ is denoted as $S^t$ and is visible to all agents.

\subsubsection{Observation.} In addition to capturing global information $S^t$, each agent needs to make a masking policy decision based on its own local observation. In general, the observation in MARL tasks is often a part of the global state. Considering these factors, we take the deep feature ViT$(P_i)$ of patch $P_i$ as the current observation value $O_i^t$ for the i-th agent. The feature extractor for local features is the same as the one for global features, both using the ViT backbone.

\subsubsection{Action.}  The action of the i-th agent aims to output whether patch $P_i$ needs to be masked. We define the action of the i-th agent as a vector $A_i$. The joint action space can be represented as $A = \{A_1, A_2, ..., A_N\}$, where $N$ represents the total number of patches. Since the action space only has two possibilities, masking or keeping the original voxels, the dimension of $A_i$ is 2. Given the current state $S^t$ and observation $O_i^t$, each agent $i$ will determine an action $a_i(S^t, O^t_i) \in A_i$ based on the current policy. The final output is the global joint action $\boldsymbol{a^t} = \{a^t_1\cup a^t_2,..., \cup  a^t_N\}$. After all patches have taken their corresponding actions through the decision policy, the time step is updated to $t = t+1$ and we obtain the enhanced volume through the decision module.

\subsubsection{Rewards.} Rewards are introduced in our MARL decision-making process in order to guide the agents to learn expected behaviors that will improve the main task's performance through more reasonable masking ratios and masking strategies, allowing the target net to better learn semantic information in the volume. Previous works \cite{lin2021selectaugment,zhang2019adversarial} attempted to increase the training loss of the target network based on rewards in order to generate deeper, more difficult-to-learn features. Inspired by these works, we refine the reward design based on the MAE pretraining paradigm. By comparing the loss difference between the data obtained from the masked data $\mathbf{x} \boldsymbol{\cdot a^{t-1}}$ generated by the previous time step's decision module and the data obtained from the current decision module $\mathbf{x} \boldsymbol{\cdot a^{t}}$, we compute the reward for the MARL policy. This encourages higher training loss during the MARL decision-making process, as shown by the equation
\begin{equation}
r^t=\mathcal{L}_{pretrain}(\phi(\mathbf{x} \boldsymbol{\cdot a^{t}}))-\mathcal{L}_{pretrain}(\phi(\mathbf{x} \boldsymbol{\cdot a^{t-1}})).
\label{rewards}
\end{equation}
In the above equation, $\mathcal{L}_{pretrain}$ represents the reconstruction loss generated by the target network, and $\phi$ denotes the target network. The accumulated reward of one sequence is
\begin{equation}
R^t=\sum_{i=t-T+1}^t \gamma^{i-1} \Bar{r}^{i},
\label{longtermrewards}
\end{equation}
where $T$ represents the desired time step length to be calculated, the discount factor $\gamma$ takes a value in (0, 1], and $\Bar{r}^{t}$ is the mean rewards at time $t$.

\subsubsection{Policy Learning.} 
\begin{figure}[t]
    \centering
    \includegraphics[width = 0.9\linewidth]{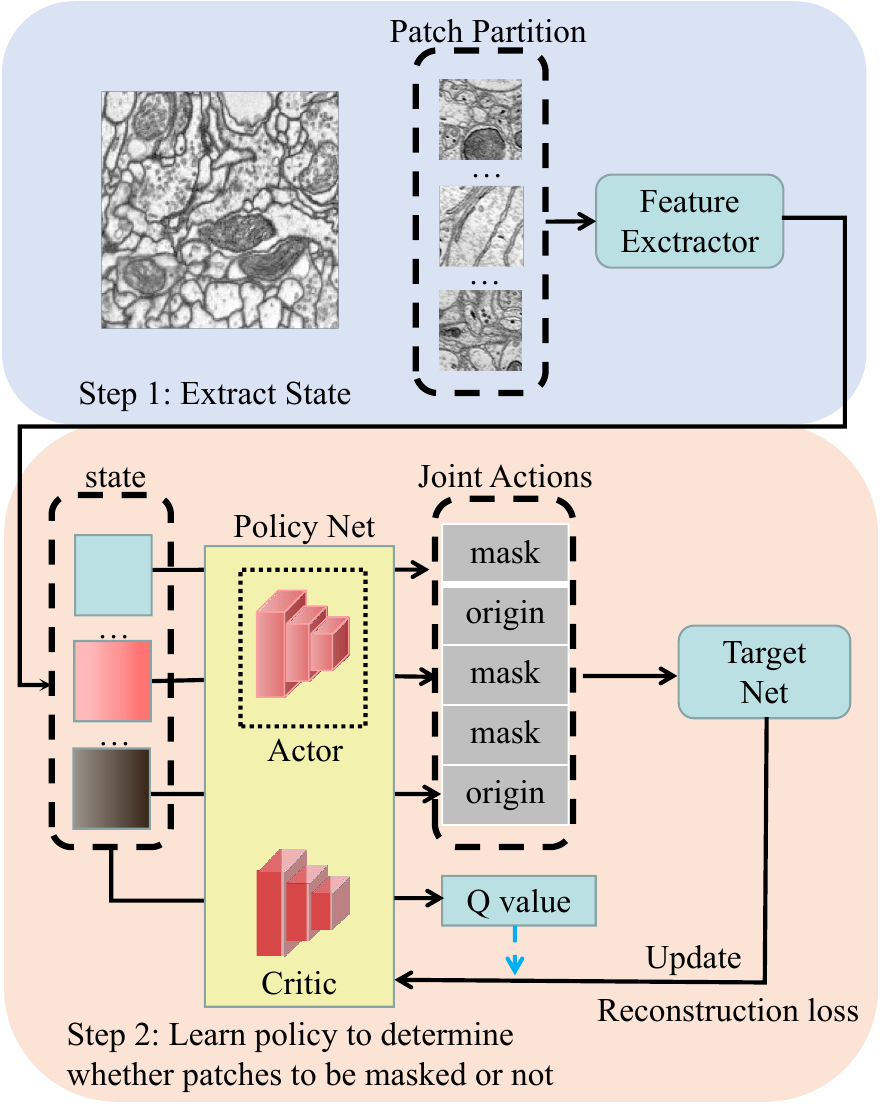}
    \caption{The framework of our proposed decision module. The first step in the process is to extract features from the encoder of the target net. In the second step, the policy network is used to decide whether or not the current patch needs to be masked. The output of the joint action constitutes the final decision.}
    \label{MARL}
\end{figure}

Considering that the action space for MARL decisions is discrete, we utilize the widely used Asynchronous Advantage Actor-Critic (A2C) algorithm \cite{mnih2016asynchronous} to perform MARL policy learning. Since the search space for actions is not large, we use simple convolution and pooling modules to adjust the Actor and Critic networks to the structure shown in Figure \ref{MARL}. The policy network is divided into an Actor and a Critic, which are adapted to the RL training algorithm. The Actor network learns a discrete control policy $\pi(a^t_i|S^t, O^t_i)$, while the Critic network aims to estimate the value of the state $V_\pi(S^t)$. We model the centralized action value function $Q$, which takes in the state information $S$ and the actions of all agents, and outputs a $Q$ value for the team, given by 
\begin{equation}
Q^{\boldsymbol{\pi}}(S^t, \boldsymbol{a^t})=E_{\boldsymbol{\pi}}\left[R_t \mid S^t, a^t_1, \cdots, a^t_N\right],
\end{equation}
where $\boldsymbol{a^t}$ represents the joint action of all agents, defined as $\boldsymbol{a}=\{a_i, \cdots, a_N\}$, and $R_t$ is the long-term discounted reward, given by equation \ref{longtermrewards}. The advantage function on the policy is then given by
\begin{equation}
A^\pi(S^t, \boldsymbol{a^t})=Q^\pi(S^t, \boldsymbol{a^t})-V^\pi(S^t),
\end{equation}
where $A^\pi(S^t, \boldsymbol{a^t})$ is the advantage of taking action $\boldsymbol{a^t}$ given state $S^t$ at time step $t$, $V^\pi(S^t)$ is the current state estimate output by Critic. It indicates that the actual accumulated reward is independent of the state and reduces the variance of the gradient. We use $\theta_p$ and $\theta_v$ to denote the parameters of the Actor and Critic, respectively. The squared value of the dominance function $A^\pi$ is taken as the loss function to update $\theta_v$ as
\begin{equation}
\mathcal{L}(\theta_v)=A^\pi(S^t, \boldsymbol{a^t})^2.
\end{equation}

To further achieve cooperative capability, the loss function of the updated Actor $\theta_p$ is defined as
\begin{equation}
\mathcal{L}(\theta_p)=-\log \pi_\theta(\boldsymbol{a^t} \mid S^t) A^\pi(S^t, \boldsymbol{a^t}),
\end{equation}
where $\pi_\theta(\boldsymbol{a^t} \mid S^t)$ is the Actor output, that is, the probability of taking each action $a^t_i$. The Actor and Critic are jointly trained in an end-to-end manner.

\subsection{Neuron Instance Segmentation Method}
The UNETR model is specifically designed for 3D segmentation tasks, such as organ segmentation. It is built upon the pretrained ViT encoder from decision-based MIM and a randomly-initialized convolutional decoder. The UNETR architecture is inspired by the U-Net model \cite{ronneberger2015u}, with the inclusion of skip connections between the encoder and decoder at multiple resolutions. The input to the UNETR decoder is a sequence of representations, denoted as ${z_3, z_6, z_9, z_{12}}$, which are reshaped to restore the spatial dimension of $\frac{H}{P} \times \frac{W}{P} \times \frac{D}{P} \times K$, where $K$ is the feature channel.

Starting from the deepest feature, $z_{12}$, each representation is processed through a varying number of deconvolutional blocks to increase its resolution by a specific factor. For example, $z_{12}$ and $z_9$ are upsampled by a factor of $\times 2$, $z_6$ is upsampled by a factor of $\times 4$, and $z_3$ is upsampled by a factor of $\times 8$. Then, representations at the same spatial resolution, such as $z_{12}$ and $z_9$, are concatenated and further upsampled to match the shape of a shallower feature. This process of concatenation and upsampling is repeated until the full resolution of the input is restored. Finally, the output layer combines the upsampled feature and the original full-resolution input to predict the segmentation map.

In this paper, for the segmentation task of EM neurons, we use a combination of affinity map-based and post-processing approaches. We first create an affinity graph based on the voxel affinity. This graph serves as the foundation for our post-processing techniques, which utilize both waterz \cite{funke2018large} and LMC \cite{beier2017multicut} to cluster the affinity map and produce the final neuron segmentation results.
Affinity-based methods have proven to be highly effective in accurately segmenting and analyzing these complex structures.

\section{Experiments}
\subsection{Training Strategy}
Our strategy consists of two phases: pretraining and fine-tuning. In the pretraining phase, to improve the training efficiency of the framework, we first pretrain decision-based MIM for 100k iterations, synchronously updating the parameters of MAE and the policy network in the decision module, and then fix the parameters of policy network and only update the parameters of MAE for another 100k iterations. In the fine-tuning phase, we load the pretrained ViT weights into the model for the downstream task and train for 200k iterations.

We use the Adam optimizer in both the pretraining and fine-tuning phases, with $\beta_1=0.9, \beta_2=0.999$. The only difference lies in the pretraining process, where we set the learning rate to 0.0001 and perform batch size 16 pretraining on 8 RTX 3090s. In the fine-tuning phase, we adopt a Layer-wise Learning Rate Decay (LLRD) training method, which adjusts the learning rate layer by layer during training. We set the learning rate of the last layer's parameters to 0.001 and the learning rate of the previous layer's parameters to 0.95 times the learning rate of the next layer's parameters. We conduct batch size 8 fine-tuning on 2 RTX 3090s. 
\subsection{Datasets and Evaluation Metrics}
\begin{table*}[ht]
\small
\centering
\renewcommand{\arraystretch}{1.1}
\scalebox{0.85}{\begin{tabular}{llcccc|cccc}
\toprule[1.2pt]
                          &                          & \multicolumn{4}{c|}{waterz \cite{funke2018large}}                                                               & \multicolumn{4}{c}{LMC \cite{beier2017multicut}}                                                                   \\ \cline{3-10} 
\multirow{-2}{*}{Dataset} & \multirow{-2}{*}{Method} & VOI$_S$ $\downarrow$      & VOI$_M$ $\downarrow$     & VOI $\downarrow$            & ARAND $\downarrow$         & VOI$_S$ $\downarrow$     & VOI$_M$ $\downarrow$     & VOI $\downarrow$            & ARAND $\downarrow$          \\ \midrule[0.8pt]
                          & superhuman \cite{lee2017superhuman}               & 0.443          & \textbf{0.320} & \cellcolor[HTML]{EFEFEF}0.763          & 0.132          & 0.578          & 0.272          & \cellcolor[HTML]{EFEFEF}0.850          & \textbf{0.131} \\
                          & MALA \cite{funke2018large}                     & 0.478          & 0.627          & \cellcolor[HTML]{EFEFEF}1.105          & 0.459          & 0.574          & 0.303          & \cellcolor[HTML]{EFEFEF}0.878          & 0.146          \\
                          & UNETR \cite{hatamizadeh2022unetr}                    & 0.495          & 0.359          & \cellcolor[HTML]{EFEFEF}0.854          & 0.153          & 0.630          & 0.278          & \cellcolor[HTML]{EFEFEF}0.908          & 0.141          \\
                          & MAE \cite{he2022masked}+UNETR                & 0.463          & 0.324          & \cellcolor[HTML]{EFEFEF}0.787          & 0.144          & 0.563          & 0.268          & \cellcolor[HTML]{EFEFEF}0.831          & 0.139          \\
                          & Dino \cite{caron2021emerging}+UNETR               & 0.482          & 0.351          & \cellcolor[HTML]{EFEFEF}0.833          & 0.150          & 0.580          & 0.287          & \cellcolor[HTML]{EFEFEF}0.867          & 0.147          \\
\multirow{-6}{*}{CREMI A}  & ours+UNETR               & \textbf{0.411} & 0.331          & \cellcolor[HTML]{EFEFEF}\textbf{0.743} & \textbf{0.131} & \textbf{0.537} & \textbf{0.260} & \cellcolor[HTML]{EFEFEF}\textbf{0.797} & 0.134          \\ \midrule[0.8pt]
                          & superhuman \cite{lee2017superhuman}               & 0.668          & 0.409          & \cellcolor[HTML]{EFEFEF}1.076          & \textbf{0.082} & 0.959          & 0.232          & \cellcolor[HTML]{EFEFEF}1.191          & \textbf{0.060} \\
                          & MALA \cite{funke2018large}                     & 0.797          & 0.512          & \cellcolor[HTML]{EFEFEF}1.309          & 0.147          & 1.060          & 0.264          & \cellcolor[HTML]{EFEFEF}1.324          & 0.084          \\
                          & UNETR \cite{hatamizadeh2022unetr}                    & 0.937          & 0.397          & \cellcolor[HTML]{EFEFEF}1.333          & 0.103          & 1.194          & 0.230          & \cellcolor[HTML]{EFEFEF}1.424          & 0.116          \\
                          & MAE \cite{he2022masked}+UNETR                & 0.776          & 0.391          & \cellcolor[HTML]{EFEFEF}1.167          & 0.099          & 0.994          & 0.224          & \cellcolor[HTML]{EFEFEF}1.218          & 0.102          \\
                          & Dino \cite{caron2021emerging}+UNETR               & 0.916          & 0.396          & \cellcolor[HTML]{EFEFEF}1.312          & 0.104          & 1.167          & 0.229          & \cellcolor[HTML]{EFEFEF}1.396          & 0.111          \\
\multirow{-6}{*}{CREMI B}  & ours+UNETR               & \textbf{0.642} & \textbf{0.381} & \cellcolor[HTML]{EFEFEF}\textbf{1.023} & 0.092          & \textbf{0.893} & \textbf{0.220} & \cellcolor[HTML]{EFEFEF}\textbf{1.113} & 0.097          \\ \midrule[0.8pt]
                          & superhuman \cite{lee2017superhuman}               & 0.943          & 0.385          & \cellcolor[HTML]{EFEFEF}1.328          & 0.134          & 1.176          & 0.260          & \cellcolor[HTML]{EFEFEF}1.436          & 0.125          \\
                          & MALA \cite{funke2018large}                     & \textbf{0.901} & 0.621          & \cellcolor[HTML]{EFEFEF}1.522          & 0.169          & \textbf{1.137} & 0.289          & \cellcolor[HTML]{EFEFEF}1.426          & 0.127          \\
                          & UNETR \cite{hatamizadeh2022unetr}                    & 0.996          & 0.423          & \cellcolor[HTML]{EFEFEF}1.419          & 0.158          & 1.417          & 0.236          & \cellcolor[HTML]{EFEFEF}1.653          & 0.148          \\
                          & MAE \cite{he2022masked}+UNETR                & 1.001          & 0.298          & \cellcolor[HTML]{EFEFEF}1.299          & 0.120          & 1.272          & 0.214          & \cellcolor[HTML]{EFEFEF}1.486          & 0.113          \\
                          & Dino \cite{caron2021emerging}+UNETR               & 1.011          & 0.412          & \cellcolor[HTML]{EFEFEF}1.423          & 0.156          & 1.364          & 0.234          & \cellcolor[HTML]{EFEFEF}1.598          & 0.146          \\
\multirow{-6}{*}{CREMI C}  & ours+UNETR               & 0.925          & \textbf{0.276} & \cellcolor[HTML]{EFEFEF}\textbf{1.201} & \textbf{0.107} & 1.194          & \textbf{0.204} & \cellcolor[HTML]{EFEFEF}\textbf{1.398} & \textbf{0.112} \\ \midrule[0.8pt]
                          & superhuman \cite{lee2017superhuman}               & 0.721          & 0.295          & \cellcolor[HTML]{EFEFEF}1.016          & \textbf{0.187} & \textbf{0.770} & 0.343          & \cellcolor[HTML]{EFEFEF}1.113          & 0.110          \\
                          & MALA \cite{funke2018large}                     & 0.734          & 0.385          & \cellcolor[HTML]{EFEFEF}1.119          & 0.305          & 0.832          & 0.357          & \cellcolor[HTML]{EFEFEF}1.189          & \textbf{0.108} \\
                          & UNETR \cite{hatamizadeh2022unetr}                    & 0.908          & 0.337          & \cellcolor[HTML]{EFEFEF}1.245          & 0.316          & 1.007          & 0.340          & \cellcolor[HTML]{EFEFEF}1.347          & 0.134          \\
                          & MAE \cite{he2022masked}+UNETR                & 0.791          & 0.306          & \cellcolor[HTML]{EFEFEF}1.097          & 0.254          & 0.888          & 0.298          & \cellcolor[HTML]{EFEFEF}1.186          & 0.120          \\
                          & Dino \cite{caron2021emerging}+UNETR               & 0.889          & 0.329          & \cellcolor[HTML]{EFEFEF}1.218          & 0.298          & 1.001          & 0.314          & \cellcolor[HTML]{EFEFEF}1.315          & 0.129          \\
\multirow{-6}{*}{AC4}     & ours+UNETR               & \textbf{0.647} & \textbf{0.285} & \cellcolor[HTML]{EFEFEF}\textbf{0.931} & 0.243          & 0.795          & \textbf{0.284} & \cellcolor[HTML]{EFEFEF}\textbf{1.079} & 0.113          \\ \bottomrule[1.2pt]
\end{tabular}}
\caption{Results on the CREMI dataset, VOI$_S$ represents split error, VOI$_M$ represents merge error, and VOI is the sum of the two. The final segmentation results are generated by using two classic post-processing methods, waterz and LMC.}
\label{cremi}
\end{table*}

\paragraph{FAFB.} The Full Adult Fly Brain (FAFB) dataset \cite{zheng2018complete} is a highly valuable resource for neuroinformatics research, offering a comprehensive and detailed view of the neural architecture of the \textit{Drosophila melanogaster} (fruit fly) brain. With a size of approximately 40 terabytes, this dataset features high-resolution 3D images with a resolution of approximately 4 nanometers per pixel, as well as manually annotated segmentation data identifying various brain structures such as neurons, glial cells, blood vessels, synapses, and other neuropil regions. In our work, we utilize the FAFB dataset as a key component in our pretraining process. To optimize the efficiency of our model, we first downsample the original dataset by a factor of 4, carefully curating a selection of 60G images that exhibit exceptional imaging quality from the entire dataset. This strategic selection ensures that our model is trained on the most accurate and reliable data possible, setting the foundation for its future performance.

\paragraph{CREMI.} The CREMI dataset is derived from the FAFB dataset, which has three manually labeled subvolumes from \textit{drosophila} brain, of which CREMI A has more regular forms, CREMI C has higher size disparities, and CREMI B has in-between segmentation difficulty. Each sub-volume has 125 slices of 1250$\times$1250 images, and we choose the first 60 slices for training, 15 slices for validation, and the remaining 50 slices for testing, which are utilized to validate our method's performance on segmentation tasks of varying difficulty.

\paragraph{AC3/AC4.} AC3/AC4 \cite{kasthuri2015saturated} are mouse somatosensory cortex datasets with 256 and 100 successive EM images (1024$\times$1024), respectively. The first 80 slices of AC3 are used as the training set, the following 20 slices as the validation set, and the first 50 slices of AC4 are used as the testing set.

\paragraph{Evaluation Metrics.} We are more interested in the model's performance in the downstream task in the self-supervised training. To assess the influence of segmentation on EM neurons, we primarily use Variation of Information (VOI) \cite{nunez2013machine} and Adapted Rand Error (Arand) \cite{arganda2015crowdsourcing} metrics. Smaller VOI and ARAND values represent better segmentation results.

\subsection{Experimental Results}
\begin{figure}[tb]
    \centering
    \includegraphics[width = \linewidth]{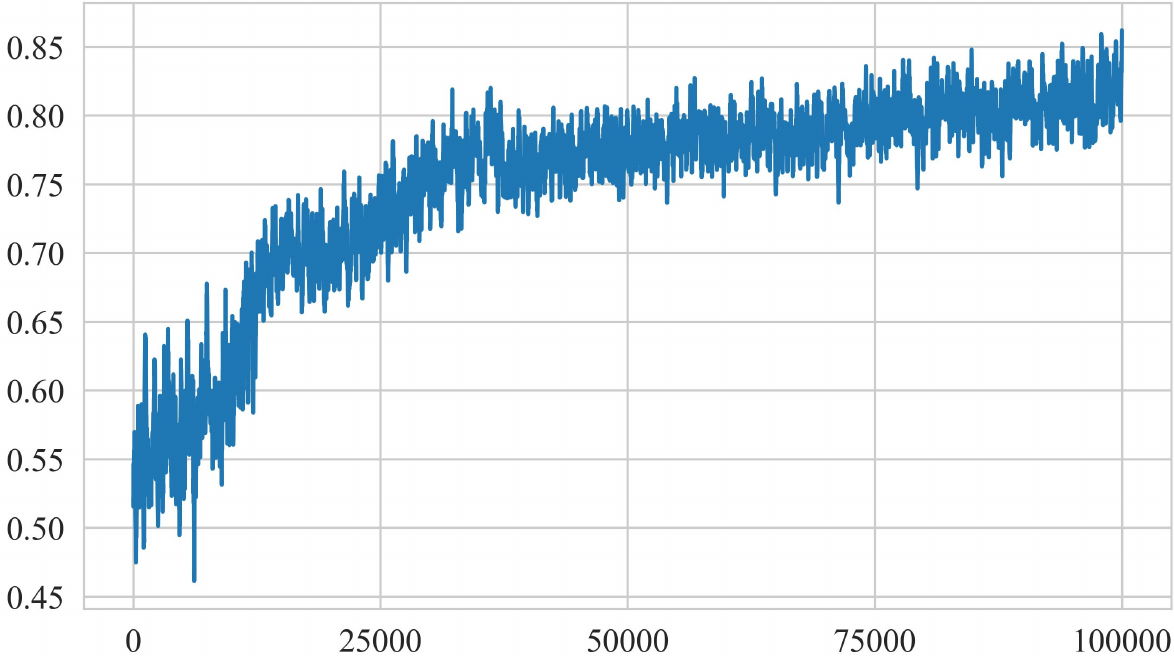}
    \caption{The variation of the masking ratio in the MARL decision process, where the final convergence result shows that the optimal masking ratio fluctuates around 0.83.}
    \label{maskratio}
\end{figure}

\paragraph{Decision Making Process.}
Decision-based MIM is pretrained using the FAFB dataset. During pretraining, the policy network in the decision module is also updated, and the real-time change in masking ratio is recorded as in Figure \ref{maskratio}. The decision-making process starts with random decisions, resulting in an overall masking ratio of around 0.5. Then, using the reconstruction loss as a reward, the decision-making principles of the agents are updated. After 50k iterations, the decision module converges and produces better reconstruction results with less lost information. During pretraining, it is found that starting with a lower masking ratio is more beneficial for the EM dataset, and gradually increasing the masking ratio helps the network's learning process progress from easy to difficult datasets, which is effective for both upstream reconstruction and downstream segmentation tasks.

\begin{figure*}[t]
    \centering
    \includegraphics[width =0.95 \linewidth]{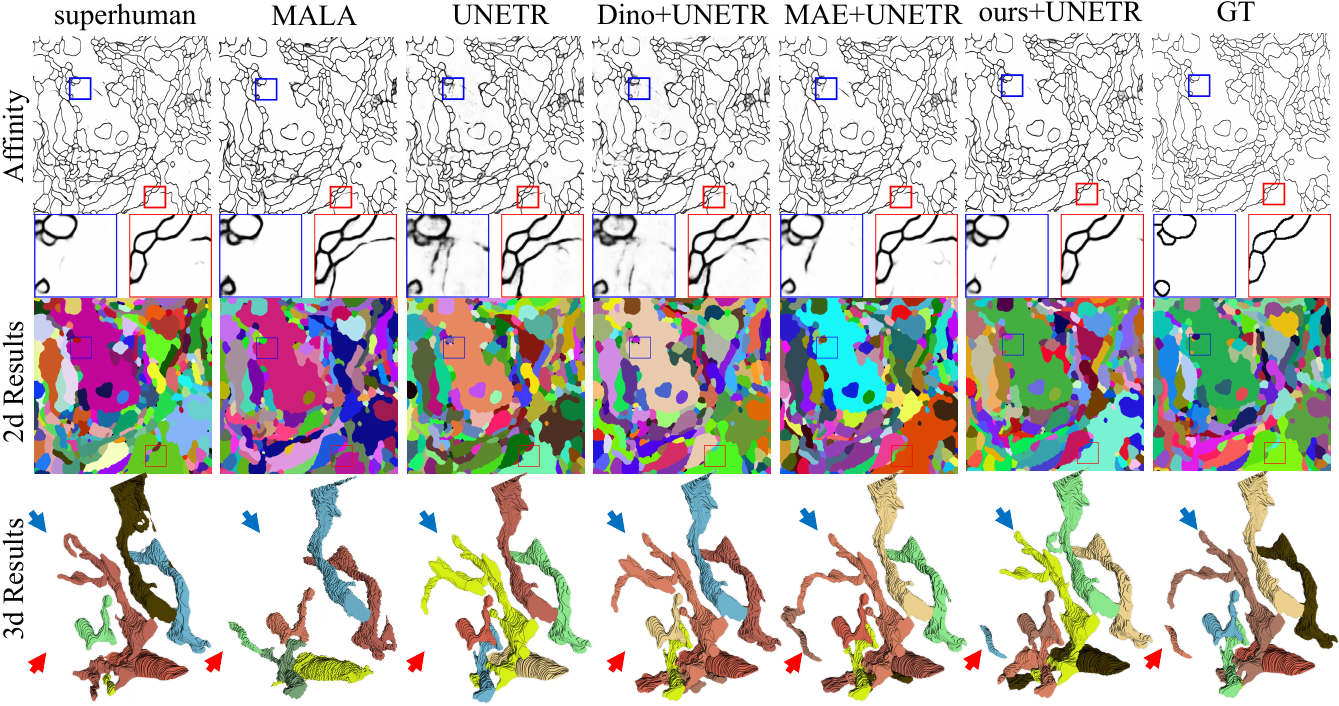}
    \caption{The visualization results in the CREMI C dataset reveal the areas of over-segmentation and under-segmentation produced by methods such as Superhuman, MALA, and UNETR.}
    \label{CREMIresults}
\end{figure*}

\paragraph{Results on CREMI.} We compare our method to the original MAE pretraining method \cite{he2022masked} and the Dino pretraining method \cite{caron2021emerging} based on contrastive learning, as well as two affinity-based Unet structures  for instance segmentation, superhuman \cite{lee2017superhuman} and MALA \cite{funke2018large}. The outcomes of the trials are provided in Table \ref{cremi} after applying both waterz and LMC post-processing procedures. Figure \ref{CREMIresults} shows the visualization results. Compared to existing self-supervised methods, our decision-based MIM approach shows a significant improvement for the downstream task and it also performs better than commonly used CNN-based methods in neuron segmentation, as can be seen in Table \ref{cremi}.

\paragraph{Results on AC3/AC4.} In order to verify the robustness and generalizability of our pretraining method, we further conduct experiments on mouse cortical neural cells. The results in Table \ref{cremi} indicate that our approach leads to a clear improvement in the downstream segmentation task, even when applied across different species.

\subsection{Ablation Study}
We use UNETR to perform a comprehensive ablation study on the AC4 dataset.

\paragraph{Effectiveness of the multi-task reconstruction.}
We conduct ablation experiments on the model's reconstruction targets. As shown in Table \ref{multitask}, it demonstrates that utilizing HOG and MSE losses together for reconstruction in the MIM outperforms using MSE or HOG alone.
\begin{table}[t]
\centering
\small
\renewcommand{\arraystretch}{1.1}
\begin{tabular}{cccc|cc}
\toprule[1.2pt]
                      &                       & \multicolumn{2}{c|}{waterz}                             & \multicolumn{2}{c}{LMC}                                 \\ \cline{3-6} 
\multirow{-2}{*}{MSE} & \multirow{-2}{*}{HOG} & VOI $\downarrow$                                    & ARAND $\downarrow$          & VOI $\downarrow$                                    & ARAND $\downarrow$          \\ \midrule[0.8pt]
$\surd$                     &                       & \cellcolor[HTML]{EFEFEF}1.097          & 0.254          & \cellcolor[HTML]{EFEFEF}1.186          & 0.120          \\
                      & $\surd$                     & \cellcolor[HTML]{EFEFEF}0.987          & 0.251          & \cellcolor[HTML]{EFEFEF}1.103          & 0.127          \\
$\surd$                     & $\surd$                     & \cellcolor[HTML]{EFEFEF}\textbf{0.957} & \textbf{0.245} & \cellcolor[HTML]{EFEFEF}\textbf{1.089} & \textbf{0.117} \\\bottomrule[1.2pt]
\end{tabular}

\caption{Ablation results of the multi-task reconstruction.}
\label{multitask}
\end{table}

\paragraph{Effectiveness of the decision module.}
We compare our proposed method to a straightforward solution that manually adjusts the masking ratio. The ablation results, as shown in Table \ref{decisionmodule}, demonstrate that our approach not only eliminates the need for manual adjustment of masking ratios but also outperforms the best results achieved through manual adjustment.
\begin{table}[t]
\centering
\small
\renewcommand{\arraystretch}{1.1}
\begin{tabular}{cccc|cc}
\toprule[1.2pt]
                             &                            & \multicolumn{2}{c|}{waterz}                             & \multicolumn{2}{c}{LMC}                                 \\ \cline{3-6} 
\multirow{-2}{*}{Rate} & \multirow{-2}{*}{Decision} & VOI $\downarrow$                                    & ARAND $\downarrow$          & VOI $\downarrow$                                    & ARAND $\downarrow$          \\ \midrule[0.8pt]
0.65                         &                            & \cellcolor[HTML]{EFEFEF}1.065          & 0.264          & \cellcolor[HTML]{EFEFEF}1.196          & 0.131          \\
0.75                         &                            & \cellcolor[HTML]{EFEFEF}0.997          & 0.258          & \cellcolor[HTML]{EFEFEF}1.132          & 0.129          \\
0.85                         &                            & \cellcolor[HTML]{EFEFEF}0.957          & 0.245          & \cellcolor[HTML]{EFEFEF}1.097          & 0.121          \\
0.95                         &                            & \cellcolor[HTML]{EFEFEF}1.005          & 0.260          & \cellcolor[HTML]{EFEFEF}1.121          & 0.127          \\
/                            & $\surd$                       & \cellcolor[HTML]{EFEFEF}\textbf{0.931} & \textbf{0.243} & \cellcolor[HTML]{EFEFEF}\textbf{1.079} & \textbf{0.113} \\ \bottomrule[1.2pt]
\end{tabular}
\caption{Ablation results of the decision module.}
\label{decisionmodule}
\end{table}
\vspace{-0.3cm}
\section{Conclusion}

In this paper, we propose a decision-based MIM approach for neuron segmentation. Our method eliminates the need for manual adjustment of masking ratios and masking strategies, using multi-agent cooperation to search for the optimal solution. Additionally, during the pretraining process, we incorporate a multi-task reconstruction and utilize HOG features to enhance the model's learning ability. Our method is validated on a variety of EM neuron datasets to demonstrate its generalizability.

\section*{Acknowledgements}
This work was supported in part by the National Natural Science Foundation of China under Grant 62021001.
\clearpage

\newpage
\appendix



\bibliographystyle{named}

\newpage

\end{document}